\pdfoutput=1

\documentclass[11pt]{article}

\usepackage{main}

\usepackage{times}
\usepackage{latexsym}

\usepackage[T1]{fontenc}

\usepackage[utf8]{inputenc}
\usepackage{booktabs}
\usepackage{amsfonts}
\usepackage{mathtools}
\usepackage{cleveref}
\crefname{section}{§}{§§}
\Crefname{section}{§}{§§}
\usepackage{microtype}
\usepackage{graphicx}
\usepackage{inconsolata}
\usepackage[ruled,linesnumbered]{algorithm2e}
\usepackage[colorinlistoftodos]{todonotes}
\usepackage{color}
\usepackage{colortbl}
\usepackage{multirow} 
\usepackage[mathscr]{eucal}
\title{MemLong: Memory-Augmented Retrieval for Long Text Modeling}

\author{Weijie Liu$^{1}$\thanks{\; Equal Contribution.}, Zecheng Tang$^1$\footnotemark[1], Juntao Li$^{1}$\thanks{\; Corresponding author.}, Kehai Chen$^2$, Min Zhang$^{1}$ \\  
 $^{1}$School of Computer Science and Technology, Soochow University \\
 $^{2}$Harbin Institute of Technology, Shenzhen \\
 \texttt{\{wjliu,zctang\}@stu.suda.edu.cn} \\
 \texttt{\{ljt,minzhang\}@suda.edu.cn}; ~~~\texttt{chenkehai@hit.edu.cn}
 }

\begin{document}
\maketitle
\begin{abstract}
Recent advancements in Large Language Models (LLMs) have yielded remarkable success across diverse fields. However, handling long contexts remains a significant challenge for LLMs due to the quadratic time and space complexity of attention mechanisms and the growing memory consumption of the key-value cache during generation. This work introduces \textbf{MemLong}: \textbf{Mem}ory-Augmented Retrieval for \textbf{Long} Text Generation (\textbf{MemLong}, a method designed to enhance the capabilities of long-context language modeling by utilizing an external retriever for historical information retrieval. MemLong combines a non-differentiable \textit{ret-mem} module with a partially trainable decoder-only language model and introduces a fine-grained, controllable retrieval attention mechanism that leverages semantic-level relevant chunks.
Comprehensive evaluations on multiple long-context language modeling benchmarks demonstrate that MemLong consistently outperforms other state-of-the-art LLMs. More importantly, MemLong can extend the context length on a single 3090 GPU from 4k up to 80k\footnote{Our code is available at ~\url{https://github.com/Bui1dMySea/MemLong}}.
\end{abstract}
\begin{figure}[t]
    \setlength{\abovecaptionskip}{0.2cm}   
    \setlength{\belowcaptionskip}{-0.4cm}   
    \centering
    \includegraphics[width=1\columnwidth]{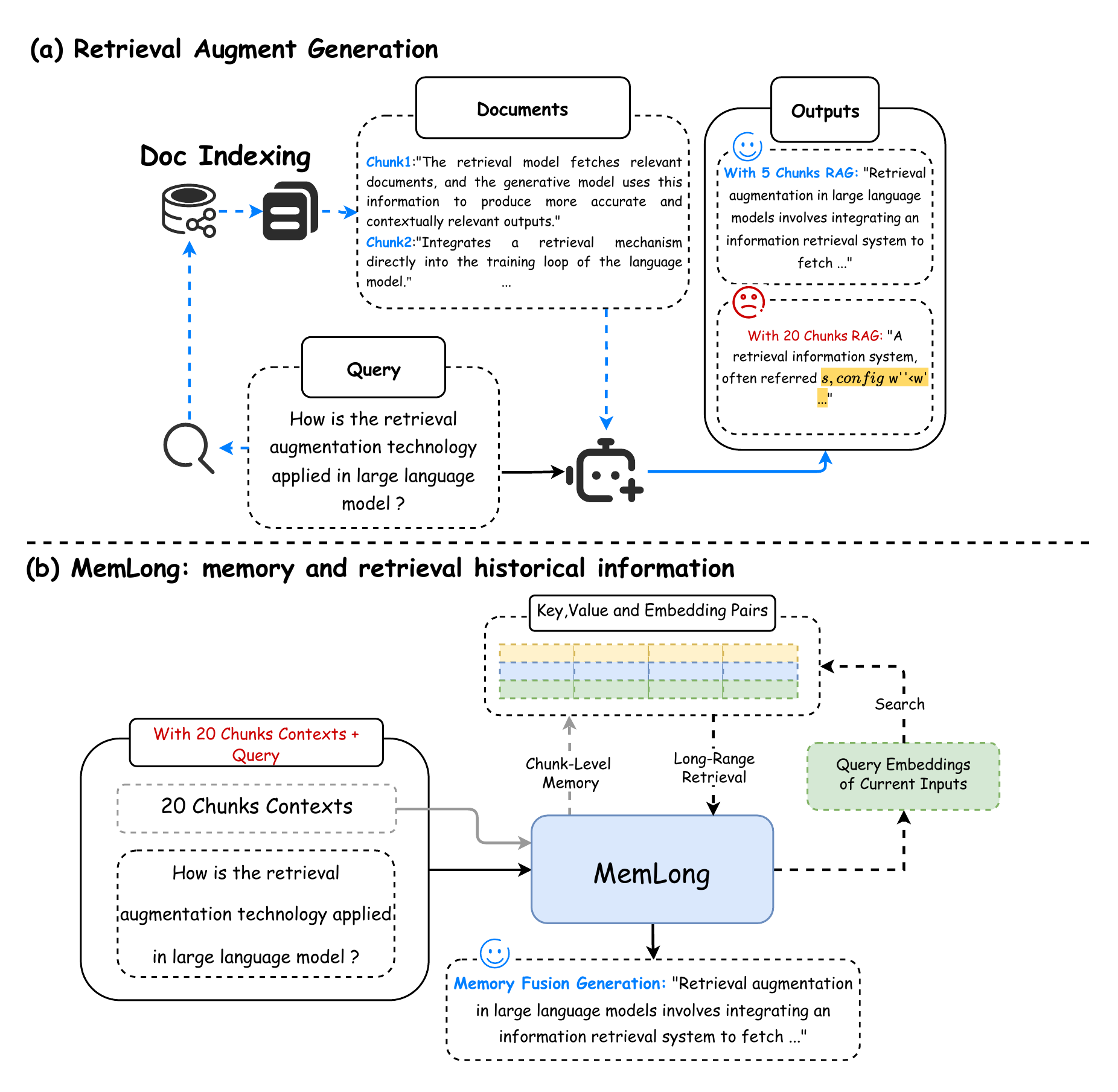}
    \caption{\textbf{Illustration of Retrieval-Augment Generation(RAG) and Memory-Retrieval flow of MemLong.} (a) RAG can even degrade the generation performance (yellow) when the length of the retrieved information exceeds the model's processing capacity. (b) Our approach utilizes an external retriever to fetch historical information, which is then passed into the model as \(\mathtt{K}\mbox{-}\mathtt{V}\) pairs rather than in text form.}
    \label{fig:1}
\end{figure}
\section{Introduction}
Large Language Models~(LLMs) have achieved remarkable success in various fields. However, due to the quadratic time and space complexity of vanilla attention mechanisms~\cite{transformer}, it is challenging to extend the context length considerably, which poses significant limitations for applications involving long-sequence tasks, such as long-document summarization~\cite{koh2022empirical} and multiple rounds of dialogue~\cite{wang2024instruct}. As a result, LLMs are often expected to maintain a long working capability~(a.k.a. long context LLMs) to effectively handle these demanding scenarios. 
\par
To tackle the computational bottleneck, numerous efforts have been made. 
The first line of work focuses on reducing the computation of vanilla attention mechanisms~\cite{transformer} by employing sparse attention operations~\cite{longformer,linformer,reformer,streamllm,longlora,longheads}. 
Although these types of works can reduce computational complexity to approximately \(\mathcal{O}(n)\), it often comes with trade-offs in model capability.
Therefore, Some works shift their focus to memory selection~\cite{transformer-xl,unlimiformer,trams}. These approaches, as token-level memory selection, can result in the truncation of semantic information.
Another recent line of work is Retrieval-Augment Language Modeling~\cite{memtrm,longmem,rpt}. These works usually introduce a retrieval mechanism to enhance the model's ability to handle long texts. However, these methods have several drawbacks. Firstly, the information stored in memory may experience \textit{distribution shifts} due to changes in model parameters during training. Secondly, these methods often require retraining, which is impractical in the era of large models. Finally, these models are often prone to processing long text inputs at the expense of the original capabilities of the pre-trained model.
To address the limitations of previous research, we posed the following question: \textbf{Can we utilize the explicit retrieval capabilities of a retriever to approximate the implicit retrieval processes within the model?}

In this work, we propose MemLong, an efficient and lightweight method to extending the context window of LLMs. \textbf{The key idea is to store past contexts and knowledge in a non-trainable memory bank and further leverages these stored embeddings to retrieve chunk-level key-value (\(\mathtt{K}\mbox{-}\mathtt{V}\)) pairs for input into the model.}. 
MemLong is applicable to any decoder-only pretrained language models by incorporating (1) an additional \textit{ret-mem} component for memory and retrieval, and (2) a \textit{retrieval causal attention} module for integrating local and memory information.
The memory and retrieval process of MemLong is illustrated in Figure~\ref{fig:1}(b). During generation,one text that exceeds the model's maximum processing length is stored as context information in a Memory Bank. Subsequently, given a recently generated text chunk in a long document, we use the retriever to explicitly retrieve past information, obtaining additional context information through index alignment. 

MemLong offers several benefits:
(1) \textbf{Distributional Consistency}:
Unlike previous models that experienced a \textit{distribution shift} when information was stored in memory, MemLong ensures the distribution of cached information remains consistent.
(2) \textbf{Training Efficient}: We freeze the \textit{lower layers} of the model and only need to finetune the \textit{upper layers} which greatly reduced computational cost. In our experiments, finetuning a 3B parameter version of MemLong on 0.5B tokens requires only eight 3090 GPUs for eight hours.
(3) \textbf{Extensive Context Window}: Since only a single layer's \(\mathtt{K}\mbox{-}\mathtt{V}\) pairs need to be memorized, MemLong is capable of extending the context window up to 80k tokens easily on a single 3090 GPU.
\par
Extensive experiments have demonstrated that MemLong exhibits superior performance in several aspects when compared with other leading LLMs. MemLong outperforms OpenLLaMA~\cite{openLLaMA} and other retrieval-based models on several long-context language modeling datasets. In retrieval-augmented in-context learning tasks, MemLong achieves an improvement of up to 10.2 percentage points over OpenLLaMA.
\section{Preliminary}
\begin{figure*}[htbp]
    \centering
    \includegraphics[width=1\linewidth]{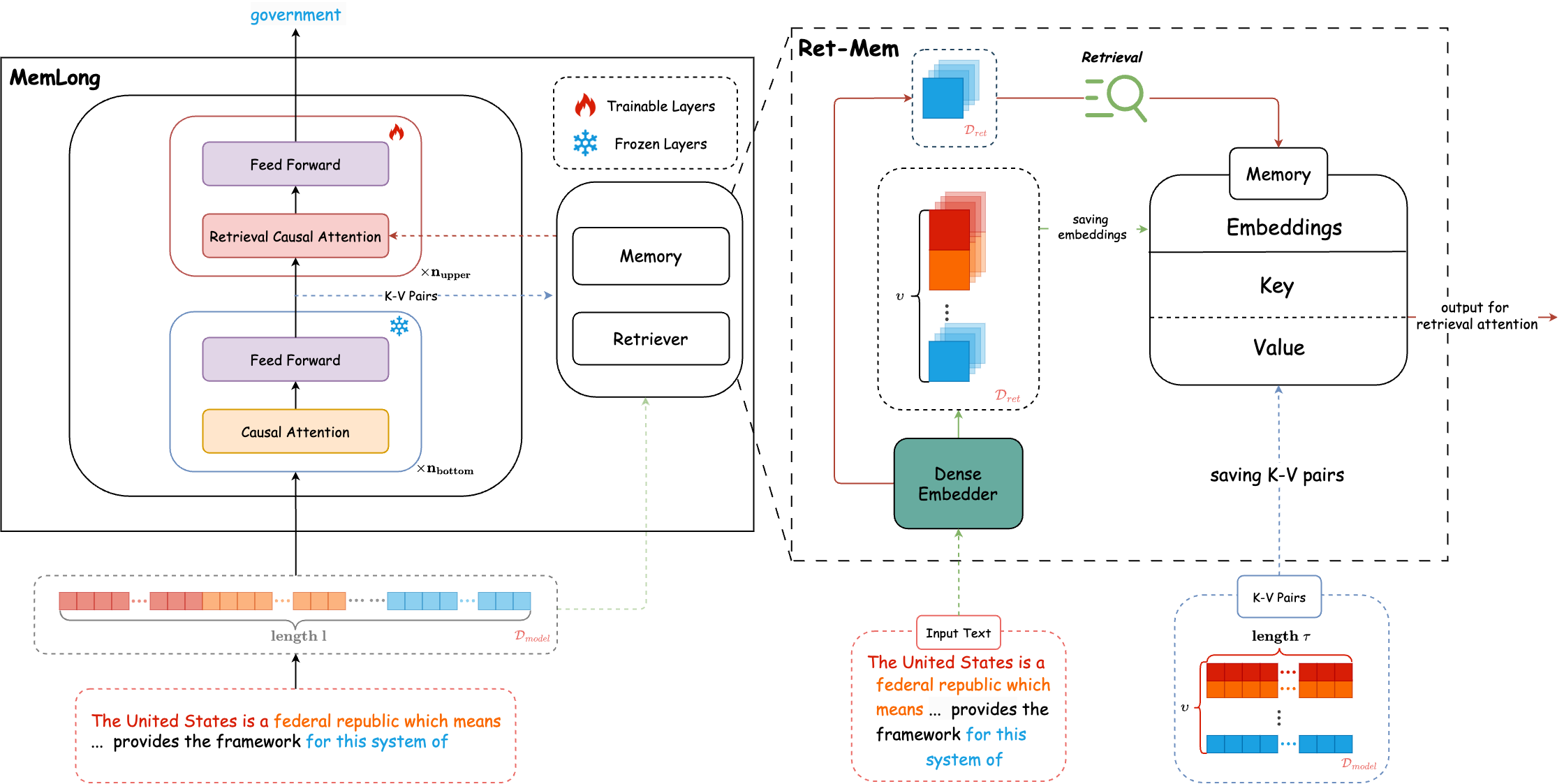}
    \caption{ An example of MemLong : In the \textit{lower layers}, where the model remains static, causal language modeling is performed on the entire chunk \(c_i\), and subsequently, \(c_i\) is cached in both embedding and \(\mathtt{K}\mbox{-}\mathtt{V}\) pair forms. Lastly, the \textit{upper layers} are finetuned to harmonize retrieval preferences and integrate the retrieved content.}
    \label{fig:2}
\end{figure*}
\subsection{Task Definition}
Language models are designed to define probability distributions over sequences of tokens, effectively predicting the likelihood of a sequence within a given language.
Given such a sequence \(x_1,\dots,x_n\), the standard approach to modeling its probability is via the next-token prediction: \(p(x_1,\dots,x_n)=\sum_{i=0}^{n}{p_{\theta}{(x_i|x_{<i})}}\), where \(x_{<i} \coloneqq x_1,\dots,x_{i-1}\) is the sequence of tokens proceeding \(x_i\). Differently from the standard language modeling objective, we not only use the current context to make next-token predictions, but also utilize external retrieval to obtain relevant information and perform knowledge fusion at the upper layers of the model.
Specifically, given a sequence consisting of \(l\) tokens and the size of each chunk \(\tau\),  we partition it into a long sequence of \( \nu = \frac{l}{\tau}\) non-overlapping chunks , which denoted as \(\mathcal{C} = (c_1,\dots,c_\nu)\). Correspondingly, its textual form is divided into \(\nu\) text chunks, which denoted as \(\mathcal{T}=(t_1,\dots,t_\nu)\). 
In each step, we perform causal language modeling on \(c_i\) in the \textit{lower layers}, while in the \textit{upper layers}, we conduct fine-grained controllable retrieval on \(t_i\) for the fusion of additional information. 
After do this, our language modeling objective becomes 
\begin{gather}
    p(x_1,\dots,x_n)=\sum_{i=0}^{n}{p_{\theta}{(x_i|\mathcal{R}(t_i),x_{<i})}}
\end{gather}
where \(\mathcal{R}(t_i)\) denotes the retrieval of \textit{neighboring chunks} of  \(t_i\)  where \(x_i\) is located.
\subsection{Module and Operation Definitions} 
As shown in Figure~\ref{fig:2}, the \textit{Ret-Mem} module comprises a \textit{Retriever} and a \textit{Memory} component for information exchange. 
Initially, we define the Memory component as \(\mathcal{M}\) and the Retriever as \(\mathcal{R}\), and their corresponding operations \(\mathcal{M}(\cdot)\) and \(\mathcal{R}(\cdot)\). Furthermore, we specify the dimension of the model as \(d_{model}\) , the dimension of the retriever as \(d_{ret}\).
The \textit{Memory} module includes two segments: \(\mathtt{K}\mbox{-}\mathtt{V}\) pairs and corresponding Representation Embeddings. The dimension for both keys and values is represented as \(\mathbb{R}^{d_{model}}\) and for Embeddings as \(\mathbb{R}^{d_{ret}}\).  It is crucial to emphasize that the actual retrieval process involves the embeddings representing the chunks, not the \(\mathtt{K}\mbox{-}\mathtt{V}\) pairs.
The \textit{Retriever} is essentially a pretrained dense embedder with excellent representation capabilities. MemLong use it to encode each chunk into Representation Embeddings. Since it produces a one-dimensional representation vector for one chunk, the memory footprint remains minimal even if the memory size is substantial.
\section{MemLong}
\subsection{Overview}
As illustrated in Figure~\ref{fig:2}, each step involves an input of a chunk \(c_i\), where the original text for that chunk  is \(t_i\).
In the \textit{lower layers} where the model is frozen, the standard causal attention is applied to the entire \(c_i\).  
For the final layer of  the \textit{lower layers}, we refer to it as the \textit{memory layer}.
Following each traversal of the \textit{memory layer}, two key operations are performed.  The first operation is retrieval, depicted by the red line, where \(t_i\) is utilized to fetch the most pertinent \(\mathtt{K}\mbox{-}\mathtt{V}\) pairs. The second operation, indicated by the blue line, involves caching the acquired \(\mathtt{K}\mbox{-}\mathtt{V}\) pairs along with their associated chunk representation.
Within the model's \textit{upper layers}, the retrieved \(\mathtt{K}\mbox{-}\mathtt{V}\) pairs are integrated with the current input context, subsequently tuning the model parameters to calibrate the retrieval reference.
Subsequent sections will explore the various facets of the MemLong framework and their intricacies, encompassing Retriever and Dynamic Memory Management~(\cref{sec:2.3}), Attention Reformulation~(\cref{sec:2.4}), and Inference with MemLong~(\cref{sec:2.5}).
\subsection{Retriever and Dynamic Memory Management}
\label{sec:2.3}
We offer a comprehensive explanation of the retrieval process and the dynamics of memory management.
\paragraph{Retrieval Process.} 
Given our objective to replace traditional kNN retrieval based on \(\mathtt{K}\mbox{-}\mathtt{V}\) pairs with explicit retrieval, we aim to pre-fetch the desired information when feasible before each model input. Specifically, for each potential query block \(c^q=c_i\) and its corresponding text block \(t^q=t_i\), we first pass it through \textit{Retriever} and then obtain a representation embedding \(r^q = \mathcal{R}(t^q)\), where \(r^q \in \mathbb{R}^{d_{ret}}\).
Subsequently, we use this representation embedding to perform retrieval against the embeddings in \(\mathcal{M}\) to obtain the required \(k\) chunk-level indices. We compute the cosine similarity between the retrieval representation \(r^q\) and the embeddings stored in Memory \(\mathcal{M}\). Finally , we get the top-k indices \(z^q = \mathtt{TopK}{\{\mathtt{Cos}\left(r^q\right)\}}\) for the \(c^q\), where \(z^q \in \mathbb{R}^{k}\).
Due to the contiguous nature within the blocks, we can easily extend the obtained indices to cover the entire relevant range for retrieval. Finally, we retrieve the corresponding \(\mathtt{K}\mbox{-}\mathtt{V}\) pairs \({\tilde{z}}^q \in \mathbb{R}^{k\times\tau\times d_{model}}\) from Memory based on these indices and used for the\textit{ upper layer}. 
It is noteworthy that we have equipped the \textit{Memory} with a counter mechanism to record the frequency of retrievals for each index contained therein. This frequency data will subsequently serve as a basis for dynamic memory updating, allowing for the prioritization of more frequently retrieved information.
\paragraph{Memory Process.}
The memory process synchronously stores the \(\mathtt{K}\mbox{-}\mathtt{V}\) pairs from the \textit{memory layer} and the \textit{representation embedding} previous calculated for retrieval , ensuring that indices for \(\mathtt{K}\mbox{-}\mathtt{V}\) pairs correspond accurately to their representation embeddings~(see Figure~\ref{fig:2}, right, blue line). For every possible chunk memory \(c^m = c_i\), and its corresponding text chunk \(t^m=t_i\), we divide the memory process into two parts: the first part details how to cache the \(\mathtt{K}\mbox{-}\mathtt{V}\) pairs, and the second part explains how to store the corresponding representations. Firstly, we input \(c^m\) into the MemLong and get the output from the \textit{memory layer}. It is worth noting that, since the lower layers are frozen during training, we can ensure that the distribution of the output \(\mathtt{K}\mbox{-}\mathtt{V}\) pairs is consistent. This consistency is crucial for avoiding the distribution shift issue, which was previously observed in models like MemTrm~\cite{memtrm}. Our memory operation is highly efficient because it only involves storing the representations needed for retrieval, \(r^m=r^q\), thereby avoiding redundancy. After the retrieval for all chunk pairs is complete, the memory operation—denoted as \(\mathcal{M}(k,v;r^m)\)—synchronously updates the memory with both the Key-Value pairs and their corresponding representations.
\paragraph{Dynamic Memory Update.} 
When memory overflows, we use the Counter to update memory intelligently. In our experiments, we keep the latest 10\% of memory content due to its potential relevance, discard the oldest 10\% as likely outdated, and prioritize the middle 80\% based on retrieval frequency, deleting the least accessed entries until memory usage drops to 50\%. This selective pruning balances recency and relevance, retaining valuable information and removing less pertinent data. Unlike traditional FIFO strategies, our method focuses on retrieval frequency to efficiently prune redundant information, maintaining a high-quality dataset. The decision to dynamically update the datastore is a trade-off between effectiveness and efficiency. For tasks requiring long-term dependencies, storing all information can enhance comprehensive processing, but for shorter-term tasks, dynamic updates are more suitable. Dynamic updates control memory size to prevent out-of-memory issues, discard stale information, and reduce retrieval overhead, ensuring efficiency without significantly compromising performance.
\subsection{Attention Reformulation}
\label{sec:2.4}
\begin{figure}[t]
    \centering
    \includegraphics[width=0.8\linewidth]{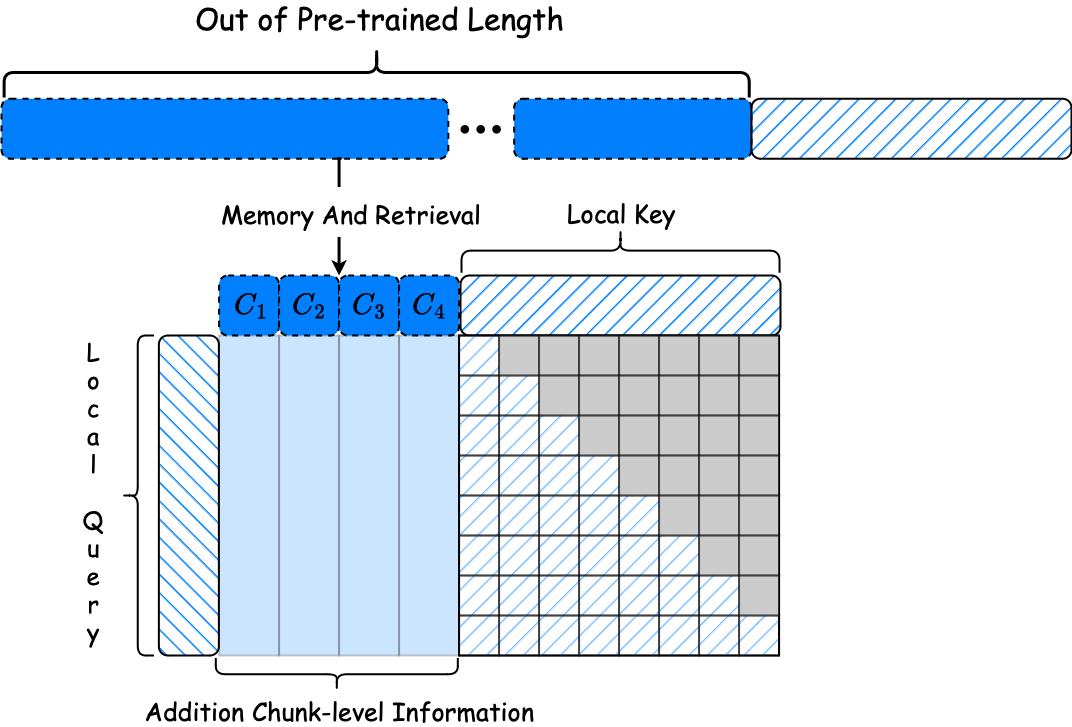}
    \caption{Illustration of retrieval causal attention. Local causal attention is applied to the recent context, while chunk-level \(\mathtt{K}\mbox{-}\mathtt{V}\) pairs, obtained through the retrieval method, enable bidirectional attention without information leakage due to their historical nature.}
    \label{fig:3}
\end{figure}
In the trainable \textit{upper layers} of the model, we revised the attentions to fuse with long-term memory. 
As illustrated in Figure~\ref{fig:3}, unlike the traditional Transformer decoder layers that utilize Multi-Head Attention~\cite{transformer}, we propose a \textit{Retrieval Causal Attention} to extend it to a joint-attention mechanism and propose a long-term memory fusion process to enable each token to attend on both local contexts and chunk-level past contexts which have complete and continuous semantics. With the head-wise hidden state output from previous layer \(H^{l-1} \in \mathbb{R}^{\lvert x \rvert\times d_{model}}\) and the corresponding retrieved key-value pairs are \({\tilde{z}}^q={\{\tilde{K_i},\tilde{V_i}\}}_{i=1}^\omega \in \mathbb{R}^{k \times \tau \times d_{model}}\), the output hidden state for the next layer \(H^{l}\) is computed as:
\begin{gather}
S_a = \mathtt{Softmax}{\left(\frac{QK^T}{d}\right)} \\
S_m = \mathtt{Concat}{\left\{\mathtt{Softmax}(\mathbf{{\tilde{z}}^q_i})\right\}}_{i=1}^{\omega}
\end{gather}
To avoid the interference caused by the retrieval attention scores \(S_m\) at the initial stage of training, we adopt a multi-head attention mechanism following the approach of the LLaMA-adapter\cite{LLaMA-adapter} :
\begin{gather}
    S_l^g = {\left[(S_m)\cdot g_l;(S_a)\right]}^T
\end{gather}
Finally, we concatenate the  \(\tilde{V}\) and \(V\) to obtain \(H^l\):
\begin{gather}
    V_l = {\left[\tilde{V}_c;V_i\right]},H^l = S_l^gV_l 
\end{gather}
\subsection{Inference with MemLong}
\label{sec:2.5}
When MemLong receives an input exceeding the length, we treat it as two segments: the \textit{prefix} and the \textit{main}. 
We will separately describe the encoding of long inputs and the generation of long outputs during the inference phase. 
When MemLong receives long inputs, it first divides the \textit{prefix} into multiple non-overlapping chunks and computes the from its \textit{memory layer}, which ensures that the number of tokens involved in the attention is equal to the chunk size, which is much smaller than the length of the input. It is important to note that each chunk is interrelated (e.g., the \(t\)-th chunk needs to process the of the previous \(t-1\) chunks). 

The second step is to select the \(k\) most relevant chunks for the \textit{main} based on chunk-level retrieval representations and to obtain their key and value representations. After this, for the upper retrieval layers, the attention window for retrieval is equivalent to \(k*\tau\), which is also smaller than the input length. Finally, both length-restricted causal attention and retrieval attention is performed efficiently.

\begin{table*}
    \centering
    \renewcommand\arraystretch{1.5}
    \resizebox{1\linewidth}{!}{
    \begin{tabular}{l c c c c c c c c c c c c c c c c c c c}
        \toprule
               & \multicolumn{4}{c}{\textbf{PG19}} & & \multicolumn{4}{c}{\textbf{Proof-pile}} & & \multicolumn{4}{c}{\textbf{BookCorpus}}  & & \multicolumn{4}{c}{\textbf{Wikitext-103}}  
        \\ \cline{2-5} \cline{7-10} \cline{12-15} \cline{17-20}
        \textbf{Model}  &  1k  &   2k  &   4k  &   16k &   &   1k  &   2k  &   4k  &   16k &  &   1k  &   2k  &   4k  &   16k &  &   1k  &   2k  &   4k  &   16k  \\ 
        \midrule 
        \rowcolor{gray} 
        \multicolumn{20}{c}{\textit{7B Model}} \\
        LLaMA-2-7B &  10.82  &   10.06  &   8.92  & - &   &   3.24  &   3.40  &   2.72  & - &  &   8.73  &   7.91  &   6.99  &   - &  &   10.82 &   6.49  &   5.66  &  - \\
        LongLoRA-7B-32k & 9.76  &   9.71  &   10.37  &   7.62 &   &   3.68  &   3.35  &   3.23  &  2.60 &  &   14.99  &   12.66  &   11.66  &   6.93 &  &   7.99  &   7.83  &   8.39  &   5.47 \\
        YARN-128k-7b & 7.22  &   7.47  &   7.17  &   - &   &   3.03  &   3.29  &   2.98  &  - &  &   7.02  &   7.54  &  7.06  &   - &  &   5.71  &   6.11  &   5.71  &   - \\
        \midrule 
        \rowcolor{gray} 
        \multicolumn{20}{c}{\textit{3B Model}} \\
        OpenLLaMA-3B &  11.60  &   9.77  &   \(>10^3\)  &  - &   &   \textbf{2.96}  &   \textbf{2.70}  &   \(>10^3\)  & - &  &   \textbf{8.97}  &   8.77  &   \(>10^3\)  & - &  &   10.57  &   8.08  &   \(>10^3\)  & - \\
        LongLLaMA-3B\(^*\) &  10.59  &   10.02  &   \(>10^3\)  &   - &   &   3.55  &   3.15  &   \(>10^3\)  &   - &  &   10.70  &   9.83  &   \(>10^3\)  & - &  &   8.88  &   8.07  &   \(>10^3\)  & - \\
        LongLLaMA-3B\(^\dag\) &  10.59  &   10.25  &   9.87  &  - &   &   3.55  &   3.22  &   \textbf{2.94}  &  - &  &   10.14  &   9.62  &   9.57  &  - &  &   10.69  &  8.33  &   7.84  &  - \\
        Phi3-128k & 11.31 & 9.90 & \textbf{9.66} & - / \textit{9.65} &  & 4.25 & 3.11 & 2.77 & - / \textit{3.08} & & 11.01 & \textbf{9.22} & \textbf{8.98} & - / \textit{9.27} & & \textbf{7.54} & \textbf{7.22} & 7.01 & - / \textit{7.20} \\
        MemLong-3B\(^*\)   &  10.66  &   10.09  &   \(>10^3\)  &  - &   &   3.58  &  3.18  &   \(>10^3\)  &   - &  &  10.37   &   9.55  &   \(>10^3\)  &   - &  &   8.72  &  7.93   &   \(>10^3\)  &    - \\
        w/ 4K Memory &  10.54  &   9.95  &   9.89  &  \textbf{9.64} &   &   3.53  &   3.16  &   3.15  &   2.99 &  &   10.18  &   9.50  &   9.57  &   9.61 &  &   8.53  &   7.92 &   7.87 &   7.99 \\
        w/ 32K Memory &  \textbf{10.53}  &   \textbf{9.85}  &   \textbf{9.83}  &   9.73 &   &   3.51  &  3.15  &   3.11  &  \textbf{2.99} &  &  9.64   &  9.56  &   \textbf{9.51}  &   \textbf{9.54} &  &   8.02  &  7.58   &   \textbf{6.89}  &   \textbf{7.09} \\ 
        \bottomrule 
    \end{tabular} 
    }
    \caption{\label{tab:1}
        Sliding window perplexity of different context window extension models on PG19, Proof-pile, BookCorpus, Wikitext-103. All experiments are conducted on one 3090 24GB GPU. LongLLaMA-3B and MemLong-3B marked with \(^*\) means evaluating without Memory, and LongLLaMA-3B marked with \(^\dag\) means evaluating with infinite memory. We also evaluate MemLong with 4K/32K Memory scenarios. "- / 6.95" indicates that the model results in an Out of Memory (OOM) error on a single GPU, while on dual GPUs it yields the corresponding result.
    }
\end{table*}

\section{Experiments}
We evaluate our proposed MemLong model on various tasks that require in-memory long-context processing: (a) long-context language modeling and retrieval-augmented language modeling; (b) scalable in-context learning capable of handling a large number of demonstration examples in memory.
\subsection{Implementation Details}
\paragraph{Training Details.} We use OpenLLaMA-3B as the pre-trained backbone LLM with rotation position coding~\cite{rope}. Due to hardware constraints, we opted to train our models using the LoRA~\cite{lora} technique. The backbone LLM holds a \(L=26,H=32,d=100\) architecture. Unless specified otherwise, we use the 13-th layer as \textit{the memory layer} and the [14,18,22,26] layers as the \textit{retrieval-augment} layers. The training for retrieval-augmented adaptation iterates only on 0.5B tokens with 1024 sequence length. MemLong's trainable parameters are from 14 to 26 layers. We utilized the slimpajama dataset sampled by~\cite{LLaMA80k} as our training corpus. 
\paragraph{Position Remapping.} There are several chunk-level \(\mathtt{K}\mbox{-}\mathtt{V}\) in the \(\mathcal{M}\) retrieved for generation. Due to the uncertainty of retrieval at each step, we need to remap position embeddings to the retrieved chunks. Same as the previous work~\cite{focus_transformer}, the local context (up to 2048 tokens) receives the standard rotary positional encoding, whereas memory keys are encoded as if they had position \(0\) in the local context window. 
\subsection{Long-Context Language Modeling}
We first evaluate MemLong on long-context language modeling benchmarks to assess basic language modeling abilities. Due to the \(\mathtt{K}\mbox{-}\mathtt{V}\) cache providing sinificant background and contextual information, MemLong can retrieve relevant \(\mathtt{K}\mbox{-}\mathtt{V}\) cache quickly and make full use of it, thereby enhancing the model's in long-context modeling tasks. 
\paragraph{Datasets.} We conducte an evaluation of our model across four extensive text benchmark datasets: English-language books PG-19~\cite{pg19} and BookCorpus~\cite{bookcorpus}, Wikipedia articles Wikitext-103~\cite{wikitext103}, and mathematical papers Proof-Pile~\cite{proofpile}. The experimental results indicate a significant perplexity improvement across all datasets. Our model is tested over various lengths ranging from 1024 to 32768 tokens. Across all datasets, our model demonstrated substantial performance gains with minimal memory overhead by leveraging an external retriever and memory. 
\paragraph{Setup.}
Following~\cite{cepe}, we calculate the perplexity on the last 2048 tokens of each sequence. This experimental setup is designed to validate the influence of different retriever sizes on the overall performance of our model. For the implementation of the efficient fine-grained retrieval, we use the \textit{faiss}~\cite{faiss} toolkit to construct an exact-search index on GPU to store the \textit{Representation Embeddings} of text chunks and perform efficient retrieval. For MemLong, we split and put the tokens over \(\text{finetune-length} = 1024\) into the \(\mathcal{M}\) used for further retrieval.
\paragraph{Baselines.}
For our experiments, we employ the OpenLLaMA-3B model as our baseline. To ensure a fair comparison, we utilize an identical LoRA configuration and finetuned the models on the same amount of data from the slimpajama dataset. Additional, we compare LongLLaMA-3B~\cite{focus_transformer}, which finetuned with the Focused Transformer (FoT) method and 5B tokens.
To perform a further comprehensive comparison, we additionally test two 7B models: LLaMA-2-7B and LongLoRA-7B-32K~\cite{longlora} and two positional encoding models: Yarn-7b-128k~\cite{yarn} and Phi3-128k~\cite{phi}. 
\paragraph{Results.}
The results are shown in Table~\ref{tab:1}.
We employ Perplexity~(PPL) as the evaluation metric for the language model. Lower PPL indicates stronger language modeling capabilities.
Compared to the two fully fine-tuned models, OpenLLaMA-3B and LLaMA-2-7B, our model demonstrates comparable performance across multiple datasets when test lengths are within their pre-trained limits (2048 for OpenLLaMA-3B and 4096 for LLaMA-2-7B). However, once the test lengths exceed these pre-trained limits, our model continues to reduce perplexity even beyond the fine-tuning length of 1024 and the pre-trained length of 2048, showcasing its superior generalizability. In contrast, the OpenLLaMA-3B and LLaMA-2-7B models fail to generalize to inputs beyond their pre-trained lengths and exhibit significantly increased memory overhead due to the quadratic complexity of attention. 
We also compare our model with LongLoRA. Although the proposed Shifted Sparse Attention in LongLoRA significantly reduces memory usage, it also diminishes the model's performance on short texts.
In contrast, LongLLaMA, which \(\mathtt{K}\mbox{-}\mathtt{V}\) pairs can also be stored, suffers from OOM issues when test lengths become excessively long due to its infinitely growing memory usage.
Positional encoding models have strong generalization capabilities. However, the performance of such methods can only guarantee that the generation performance over long distances does not degrade.
\textbf{Compared to their methods, MemLong leverages an external retriever to handle longer input tokens and achieve better perplexity improvements.At the same time, because of the high storage efficiency, MemLong can effectively control the use of GPU to avoid OOM problems.}
\subsection{In Context Learning}
\begin{table*}
    \centering
    \renewcommand\arraystretch{0.9}
    \resizebox{0.9\linewidth}{!}{
    \begin{tabular}{l c c c c c c c c}
    \toprule
    \multirow{2}{*}{\textbf{Model}} & \textbf{In-Context} & \textbf{In-Memory} & \textbf{SST-2} & \textbf{MR} & \textbf{Subj} & \textbf{SST-5} & \textbf{MPQA} & \multirow{2}{*}{\textbf{Avg.}} \\
    & \textbf{\#Demons.} & \textbf{\#Demons.} & ACC$\uparrow$ & ACC$\uparrow$ & ACC$\uparrow$ & ACC$\uparrow$ & ACC$\uparrow$ \\
    \midrule
    OpenLLaMA & 4 & N/A & 90.7 & 84.0 & 58.2 & 41.0 & 70.5 & 68.9 \\
    \multicolumn{1}{c}{w./ Rag} & 4 & 4 & 90.9 & 90.5 & 61.6 & 39.2 & 63.2 & 69.1 \\
    LongLLaMA & 4 & 4 & 90.4 & 83.9 & 64.3 & 40.0 & 64.2 & 68.6 \\
    MemLong & 4 & 4 & 91.5 & 84.5 & 61.5 & 41.4 & 70.2 & \textbf{69.8} \\
    \midrule
    LongLLaMA & 4 & 18 & 91.4 & 87.1 & 59.1 & 41.0 & 64.5 & 68.7 \\
    MemLong & 4 & 18 & 91.0 & 89.6 & 61.7 & 43.5 & 69.4  & \textbf{71.0}  \\
    \midrule
    OpenLLaMA & 20 & N/A & 93.6 & 91.2 & 55.4 & 38.2 & 66.4 & 69.0 \\
    \multicolumn{1}{c}{w./ Rag} & 20 & 18 & 92.2 & 91.3 & 75.8 & 39.8 & 57.6 & 71.3 \\
    LongLLaMA & 20 & 18 & \textbf{94.1} & 90.8 & 64.2 & 41.4 & \textbf{72.1} & 72.7 \\
    MemLong & 20 & 18 & 93.5 & \textbf{93.8} & \textbf{65.8} & \textbf{43.3} & 70.6 & \textbf{73.4} \\
    \bottomrule
\end{tabular} 
    }
    \caption{\label{tab:2}
        Accuracy [\%] of 4-shot and 20-shot ICL on 5 NLU tasks (SST-2, MR, Subj, SST-5, MPQA). We compare MemLong with both the vanilla model (OpenLLaMA) and the memory-augmented model (LongLLaMA). Across a diverse range of experimental settings, our method consistently shows competitive performance.
    }
\end{table*}

Traditional in-context learning~(ICL;~\citealp{icl}) inputs few-shot non-parameterized demonstration examples along with the query into the model. However, these methods are typically constrained by the model's input length. In this experiment, since MemLong can store examples in a parameterized form within its memory, we primarily investigate whether MemLong can effectively utilize the knowledge stored in its memory to enhance its emergent abilities.  The results are shown in Table~\ref{tab:2}. 
Compared to OpenLLaMA,which rely solely on non-parametric knowledge , given the same number of in-context demonstrations, MemLong can utilize additional demonstrations stored in its memory. The performance further increases or remains consistent with more demonstrations in the memory. 
In our comparative analysis against LongLLaMA, it was observed that our model outperforms LongLLaMA across the majority of datasets under the same conditions of preserving In-Memory Demonstrations. \textbf{It is important to highlight that our model operates with significantly lower training parameters (200M \textsc{V.\ S.} 0.3B) and fine-tuning data volume (0.5B \textsc{V.\ S.} 5B) compared to LongLLaMA.} This underscores our model's efficiency in leveraging an external retriever for information acquisition, demonstrating a superior ability to synthesize and utilize knowledge effectively with substantially fewer resources.
\section{Ablation Study}
\subsection{Training Setting}
\begin{figure}[t]
    \centering
    \includegraphics[width=1\linewidth]{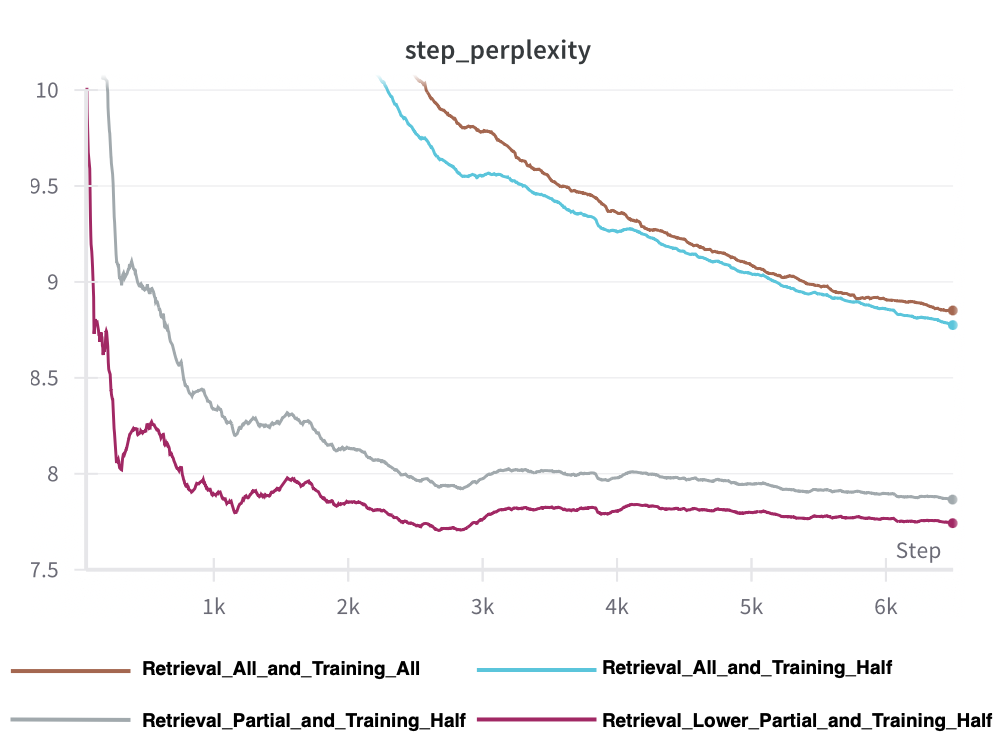}
    \caption{Degree of PPL during the training phase. The indicator for the y-axis is PPL. We mainly focus on training params and retrieval layers. We provide the specific parameter settings of each line in \cref{sec:1}.}
    \label{fig:4}
\end{figure}

During the training phase, we explore the effects of varying retrieval layers on the model and examine whether the distribution shift problem, as discussed in MemTrm~\cite{memtrm}, could be adequately resolved by our approach.
As mentioned before, Our method proposes a low-cost solution for distribution shifts. As shown in Figure~\ref{fig:4}, the brown line (the line at the top of the picture; the training method is similar to MemTrm fine-tuning all parameters of the model and all layers after the \textit{memory layer} are involved in the retrieval) is significantly worse than all other ours methods (even the most unreasonable settings) in terms of performance and fitting speed. We will analyze the performance of the reasoning stage later.  
\subsection{Inference Performance}
\begin{table}[h]
    \centering
    \renewcommand\arraystretch{1.3}
    \resizebox{1\linewidth}{!}{
    \begin{tabular}{l c c c c c c c}
    \toprule
         & \multicolumn{3}{c}{\textbf{PG19}} & & \multicolumn{3}{c}{\textbf{Proof-pile}} \\ \cline{2-4} \cline{6-8}
    \textbf{Method}    & 2k & 4k & 8k & & 2k & 4k & 8k \\ 
    \midrule
    \textbf{MemLong}\(^*\) & 10.09 & \(>10^3\) & - & & 3.18 & \(>10^3\) & - \\ 
    \midrule
    w. RA + TA  & 11.43 & 11.40 & 10.65 & & 3.51 & 3.26 & 3.14 \\ 
    w. RA + TH  & 10.57 & 10.48 & 10.36 & & 3.30  & 3.26 & 3.15 \\ 
    w. RP + TH  & 10.28 & 10.15 & 10.12 & & 3.21 & 3.13 & 3.08 \\ 
    w. RLP + TH  & \textbf{9.85} & 9.83 & \textbf{9.80} & & \textbf{3.15} & \textbf{3.11} & \textbf{3.04}\\ 
    \bottomrule
    \end{tabular}
    }
    \caption{Different retrieval layers can affect MemLong's performance. MemLong marked with \(^*\) means evaluating without Memory. The size of all methods using Memory is set to 32768. RA means retrieval across all upper layers;  TA means training all params without freeze; RP means retrieval across fewer upper layers; RPL means retrieval acorss much fewer upper layers.}
    \label{tab:3}
\end{table}
\paragraph{Q1: Does the memory length affect the performance of the model ?}
\begin{figure}[t]
    \centering
    \includegraphics[width=1\linewidth]{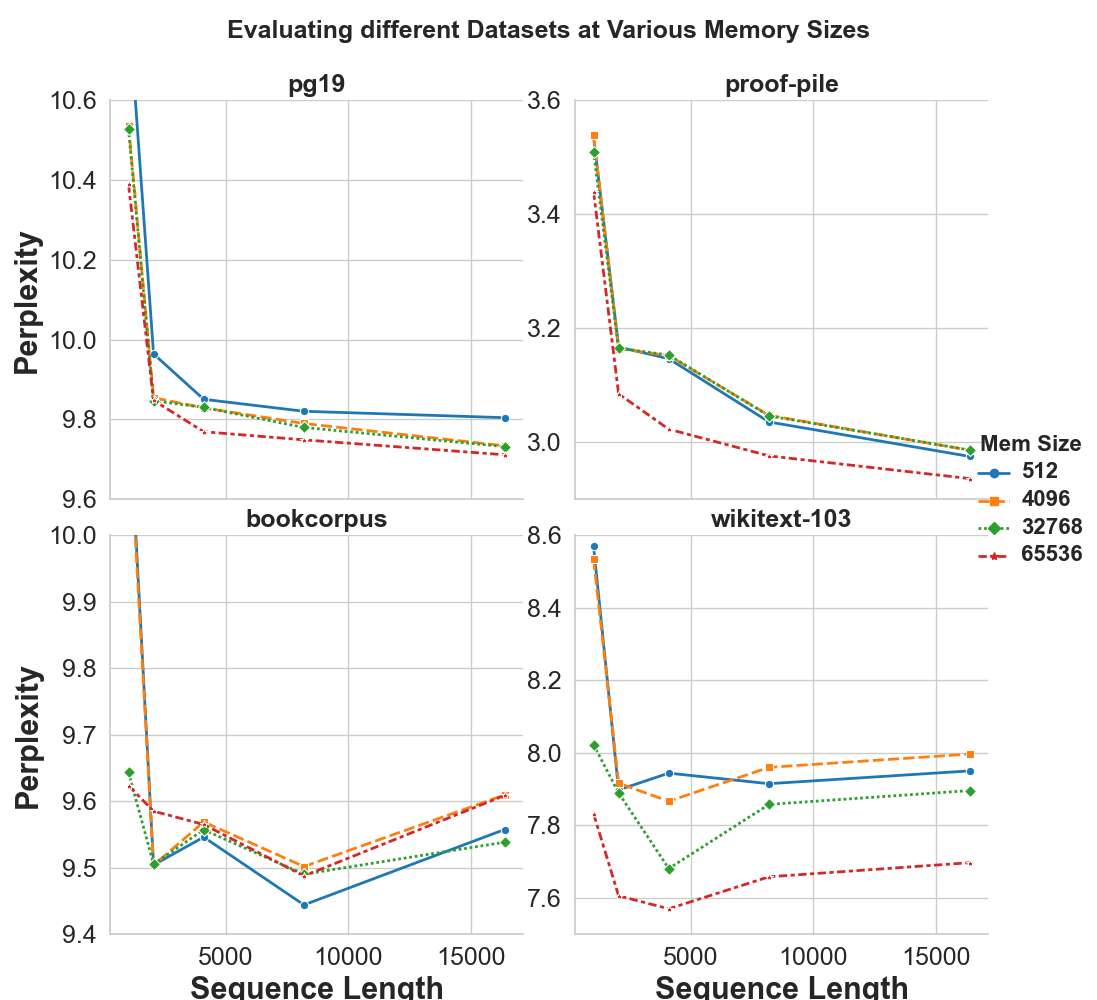}
    \caption{Evaluating different datasets at various memory sizes.In each subplot, all parameters are the same except for the memory size.}
    \label{fig:5}
    \vspace{-0.5cm}
\end{figure}
As depicted in Figure~\ref{fig:5}, our examination of the same model's performance across various memory sizes demonstrates a clear correlation between memory capacity and model efficiency. The trend indicates that incremental increases in memory size yield gradual enhancements in performance. Moreover, a critical threshold is identified at a memory size of 65536, beyond which the model's capabilities undergo a substantial leap. This suggests that while expanding memory offers substantial benefits, there is a practical ceiling to its effectiveness, likely influenced by the nuances of the data's distribution. 
\paragraph{Q2: How many layers do we need to introduce extra memory information?}
As shown in Figure~\ref{fig:4}, (the pink line) and Table~\ref{tab:3} (RPL+TH), the model performs best when the number of retrieval layers is set to [13,17,21,25]. It is empirically believed that if retrieval information is introduced into all upper layers of the model, it leads to a decrease in the model's attention to local context. Therefore, selecting retrieval layers at appropriate intervals can actually enhance the model's capabilities.
\section{Related Work}
\subsection{Long Context Language Modeling}
Long context Language Modeling mainly concentrate on length extension and context window expansion.
Length Extension studies typically target the popular RoPE encoding, aiming to scale unseen PE into the space of positions seen during pre-training. These works~\cite{rope,alibi,pi,yarn} enable the model to generalize to unseen positional encodings during inference, thereby achieving extrapolation beyond the lengths encountered during training. In contrast, our method does not require modifying the PE, and only use one addition module to extend the context. Context Window Extension focuses on how to extend the context window that LLMs can handle the input at one time. Due to the quadratic time and space complexity of computing attention, extending the input length of language models is quite challenging. Sparse attention  ~\cite{reformer,longlora,focus_transformer,unlimiformer,longformer} techniques have made significant strides, but our focus is on improving long-range language modeling by enabling LLMs to access relevant information at shorter input lengths via a retrieval-enhanced method.

\subsection{Retrieval-Augmented Language Modeling}
Much effort has been made to enhance Retrieval-Augmented Language Modeling~\cite{rag,fid,in_context_ralm,genread,self-rag}. While some approaches use external retrievers, non-parametric information fusion often falls short compared to parametric methods within the model. We concentrate on integrating retrieval concepts directly into the model. REALM~\cite{realm} suggests that relying solely on internal model knowledge is inefficient and advocates for the model to learn to retrieve and comprehend. kNN-LM~\cite{knn-lm} enhances language modeling by blending the LLM's next-word predictions with those from a retrieval-based mechanism. MemTrm~\cite{memtrm} introduces a memory bank but risks shifting memory distributions due to parameter adjustments. LongMEM~\cite{longmem} mitigates this by training a sub-network, though this adds significant overhead. In contrast, our approach involves a fixed pre-trained model, enhancing it with a frozen retriever that aligns with the model's internal retrieval processes, thus avoiding distribution shifts and architectural changes.
\section{Conclusion}
We introduce MemLong, an innovative approach that significantly enhances the capability of language models to process long texts by leveraging an external retriever. MemLong utilizes a proficient retriever to swiftly and accurately access text relevant to the distant context with minimal memory overhead. MemLong successfully expands the model's context window from 2k to 80k tokens. We demonstrate that MemLong exhibits considerable competitive advantages in long-distance text modeling and comprehension tasks. MemLong can achieve up to a 10.4 percentage point improvement in performance compared to the full-context model.
\section*{Limitations}
Our work primarily focuses on OpenLLaMA-3B. We hope that future research will explore and investigate the application of our methods to models of various sizes. At the same time, it has been found that while single-layer K-V Pairs can provide additional semantic information to the upper layers, this information is unstable. We hope that future work can provide a more rational framework to accommodate our methods. At the same time, we employ a retriever with fixed FlagEmbeddings~\cite{bge_embedding,llm_embedder}, but studying a greater range of retrievers would be useful.
\section*{Ethics Statement}
In the pursuit of advancing knowledge and developing innovative solutions, we are committed to upholding the highest ethical standards. Our work is guided by a steadfast dedication to integrity, transparency, and respect for all individuals and communities involved. Since pre-trained models may have some bias due to the unavoidable presence of harmful/offensive corpus during training, MemLong fine-tuning on Slimpajama will face this problem as well. Although solving this problem is out of our current work, we hope that there will be future work that addresses this type of problem well.
\bibliography{main}
\clearpage
\appendix
\begin{table*}[htbp]
    \centering
    \renewcommand\arraystretch{1}
    \resizebox{1\linewidth}{!}{
    \begin{tabular}{l|c|c|c}
    \toprule
           Setting Name  & Retreival Layers &  Memory Layer & Training Params \\
    \midrule
      Retreival\_All\_and\_Training\_All & [14,15,\(\dots\),26]  & 13 & All of Model's Trainable \\
      Retreival\_All\_and\_Training\_Half & [14,15,\(\dots\),26]  & 13 & Half of Model's Trainable \\
      Retreival\_Partial\_and\_Training\_Half & [14,16,18,\(\dots\),26]  & 13 & Half of Model's Trainable \\
      Retreival\_lower\_Partial\_and\_Training\_Half & [14,18,22,26]  & 13 & Half of Model's Trainable \\
    \bottomrule
    \end{tabular}
    }
    \caption{The specific parameters of different setting names.}
    \label{tab:4}
\end{table*}

\section{Different Training Settings}
\label{sec:1}
As shown in ~\ref{tab:4}, we list the variable values corresponding to different setting names in the ablation experiment.

\end{document}